\pdfoutput=1

\documentclass[11pt]{article}

\usepackage[final]{acl}

\usepackage{times}
\usepackage{latexsym}

\usepackage[T1]{fontenc}

\usepackage[utf8]{inputenc}

\usepackage{microtype}

\usepackage{inconsolata}

\usepackage{graphicx}

\usepackage[utf8]{inputenc} 
\usepackage[T1]{fontenc}    
\usepackage{hyperref}       
\usepackage{url}            
\usepackage{booktabs}       
\usepackage{amsfonts}       
\usepackage{nicefrac}       
\usepackage{microtype}      
\usepackage{xcolor}         

\usepackage{amsfonts}
\usepackage{soul}
\usepackage{amsmath}
\usepackage{ulem}

\usepackage{tcolorbox}
\usepackage{lipsum} 
\newcommand{\VarSty}[1]{\textnormal{\ttfamily\color{blue!90!black}#1}\unskip}

\usepackage{multirow}
\usepackage{graphicx}
\usepackage{enumitem}
\usepackage{array}

\title{What Makes for Good Image Captions?}

\author{
 \textbf{Delong Chen\textsuperscript{1,2}}\quad
 \textbf{Samuel Cahyawijaya\textsuperscript{1}}\quad
 \textbf{Etsuko Ishii\textsuperscript{1}}\quad
\\
 \textbf{Ho Shu Chan\textsuperscript{1}}\quad
 \textbf{Yejin Bang \textsuperscript{1,2}}\quad
 \textbf{Pascale Fung \textsuperscript{1,2}}
\\
\\
 \textsuperscript{1}HKUST\quad
 \textsuperscript{2}Meta FAIR Paris
}

\begin{document}

\newtheorem{theorem}{Theorem}
\newtheorem{definition}{Definition}
\newtheorem{assumption}{Assumption}
\newtheorem{proof}{Proof}

\maketitle
\begin{abstract}

    This paper establishes a formal information-theoretic framework for image captioning, conceptualizing captions as \textit{compressed linguistic representations} that selectively encode semantic units in images. Our framework posits that good image captions should balance three key aspects: informationally sufficient, minimally redundant, and readily comprehensible by humans. By formulating these aspects as quantitative measures with adjustable weights, our framework provides a flexible foundation for analyzing and optimizing image captioning systems across diverse task requirements. To demonstrate its applicability, we introduce the Pyramid of Captions (PoCa) method, which generates enriched captions by integrating local and global visual information. We present both theoretical proof that PoCa improves caption quality under certain assumptions, and empirical validation of its effectiveness across various image captioning models and datasets. 

\end{abstract}

\section{Introduction}
\label{Section 1 Introduction}

    Image captioning, the process of translating visual content into natural language descriptions, serving pivotal roles in real-world applications ranging from assisting visually impaired individuals~\cite{gurari2020captioning,rane2021image,guo2022offline,ganesan2022deep} to facilitating content-based image retrieval~\cite{gudivada1995content,srihari1995automatic,datta2005content,da2006content,jain2015content,li2021recent}. Over the last decade, the field of image captioning has witnessed substantial progress, primarily driven by advancements in deep neural nets and the availability of large-scale high-quality image-text datasets. 
    
    Despite empirical advancements, several fundamental questions remain unanswered: \textit{What makes for good image captions? Which properties should they possess, and how can we measure them?}  Some existing models can generate captions closely resembling single-sentence human annotations~\cite{ChenFLVGDZ15}, but these may not be adequate for use cases where more comprehensive coverage of fine-grained visual information is required. Conversely, recent Large Vision Language Models (LVLMs)~\cite{liu2023visual, dai2024instructblip} have demonstrated the ability to generate multi-paragraph detailed image descriptions~\cite{urbanek2023picture, cho2023davidsonian, liu2023visual, wang2024asm}. Yet, longer captions can sometimes be less accurate, hallucinate content, or put excessive emphasis on irrelevant details while omitting important ones.

    Recognizing the absence of a universal standard for ideal captions, this work aims to establish well-defined principles for image captioning that address varying task requirements. We introduce an information-theoretic framework based on semantic communication~\cite{Peyrard2019Simple, Zhong2017Theory} and the information bottleneck principle~\cite{Tishby2000information,ShwartzZiv2017Opening, Tsai2021Self}. By leveraging this perspective, we formulate an objective function for image captioning that strikes a balance among three key criteria:
    
    \begin{itemize}
        \item \textbf{Information Sufficiency}: Ensuring comprehensive coverage of meaningful content, measured by the mutual information between the caption and task-relevant visual semantics.
        \item \textbf{Minimal Redundancy}: Optimizing the conciseness of the caption, quantified by the entropy of the generated caption.
        \item \textbf{Human Comprehensibility}: Facilitating ease of understanding for human readers, assessed through the distributional distance between generated captions and natural language.
    \end{itemize}

    Our framework conceptualizes images and captions as observations of latent variables in a semantic space. This allows us to formulate image captioning as a communication process where semantic information is transmitted from the image to the caption, and measure the error at the semantic level. We then present formal quantitative measurements of the above three criteria and define the ultimate objective of image captioning as a weighted combination of them. Varying the weighting coefficients of these terms suits different preferences over image captions (\textit{e.g.}, comprehensiveness, succinctness, readability), providing a rigorous foundation for analysis and evaluations.

    Our framework provides a rigorous foundation for analyzing and advancing image captioning techniques. To demonstrate its practical applicability, we present the \textbf{Pyramid of Captions (PoCa)} method as an example application. PoCa employs a hierarchical approach to generate semantically rich captions by leveraging both local and global visual information. Utilizing our theoretical framework, we provide formal proof that each local-global aggregation operation in PoCa improves caption quality under certain assumptions. Empirical evaluations across various image captioning models and datasets corroborate our theoretical findings, showing that PoCa consistently yields more informative and semantically aligned captions while maintaining brevity and interpretability.

\section{Proposed Framework}
\label{Section 2 Framework}

\begin{figure}[h!]
    \centering
    \includegraphics[width=\columnwidth]{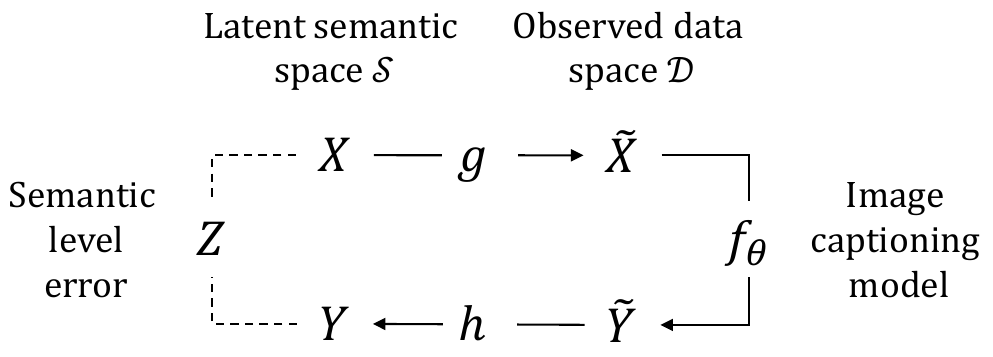}
    {\small\caption{\label{fig:framework_overview}
        Overview of our formulation. Some latent variable $X$ in a latent semantic space $\mathcal{S}$ generates image $\tilde{X}$ in data space $\mathcal{D}$. The image $\tilde{X}$ is then captioned by $f_\theta$ producing a caption $\tilde{Y}$ which can be mapped back to the original latent space as $Y$. The semantic-level error $Z=Y-X$ measures the difference between the source semantics $X$ and received semantics $Y$.
        }}
\vspace{-10pt}
\end{figure}

In this section, we provide a theoretical framework for image captioning as depicted in Figure~\ref{fig:framework_overview}. First, we formulate the task of image captioning by applying the concept of semantic units~\cite{Peyrard2019Simple, Zhong2017Theory}. We suppose that an image is an observation of a latent variable in a semantic space characterized by semantic units. An image captioning model will generate a caption for the image, and the caption can be mapped back to the latent semantic space and compared with the source semantic latent variable.

Based on this framework, we then introduce our proposed objectives inspired by the information bottleneck principle~\cite{Tishby2000information, ShwartzZiv2017Opening, Tsai2021Self} for feature representation learning~\cite{Tsai2021Self}. 
In our framework, we consider that the overall image captioning objective is composed of a \textit{``information sufficiency''} term, a \textit{``minimal redundancy''} term, and a \textit{``human comprehensibility''} term.

\subsection{Formulation}

We assume that meaning arises from a set of independent and discrete units called semantic units in a semantic space, and images and captions are observations of some latent variables in this semantic space~\cite{Peyrard2019Simple, Zhong2017Theory}. Following~\cite{Peyrard2019Simple, Zhong2017Theory}, we define a semantic unit and the semantic space as:

\begin{definition}\label{def:semantic units}
    \textbf{\upshape (Semantic Units and Semantic Space)} A semantic unit represents an atomic piece of information. The set of all possible semantic units is denoted by $\Omega = \{w_i\}_{i=1}^n$. A semantic space is an $n$-dimensional space $\mathcal{S} \in [0,1]^n$ where the value of the $i$-th dimension of represents the probability of the presence of the corresponding semantic unit $p(w_i)$.
\end{definition}


The $\Omega$ encompasses a wide range of semantic units, similar to how a vocabulary contains diverse words. Each point in $\mathcal{S}$ corresponds to a specific combination of semantic units, corresponding to different meanings. Adopting the semantic space and its probabilistic interpretation in~\cite{Peyrard2019Simple,Shao2022Theory,Niu2024Mathematical}, we can apply the classical information theory~\cite{Shannon1948mathematical} to operate at the semantic level, and make information in images and captions to be comparable. 

Let a random variable $X \in \mathcal{S}$ and $Y \in \mathcal{S}$ represent semantic information in an image and a caption, where $X_i=p(w_i)$ and $Y_i=p(w_i)$ represent the likelihood of a semantic unit $w_i$ observed in an image $\tilde{X} \in \mathcal{D_\mathrm{image}}$ or a caption $\tilde{Y} \in \mathcal{D_\mathrm{caption}}$. These latent variables $X$ and $Y$ encode all the information within real images $\tilde{X} \in \mathcal{D_\mathrm{image}}$ and textual captions $\tilde{Y} \in \mathcal{D_\mathrm{caption}}$, both low-level and high-level semantics. Then, image captioning can be framed as follows:

\begin{definition}\label{def:image captioning}
    \textbf{\upshape (Image Captioning)} An image captioning model $f$ parameterized by $\theta$ operates in the observed data spaces $f_\theta:\mathcal{D_\mathrm{image}} \rightarrow \mathcal{D_\mathrm{caption}}$, it translates an image into a caption, \textit{i.e.}, $\tilde{Y}=f_\theta(\tilde{X})$. 
    $\tilde{X}$ is generated from some source latent variable $\tilde{X}=g(X)$ while $\tilde{Y}$ can be converted back to the latent semantic space $Y=h(\tilde{Y})$. 
    Let $Z=Y-X\in[-1, 1]^n$ denote the error between the source and recovered semantics caused by parameters $\theta$.
\end{definition}

Here, $Z$ can be associated with which kind of error $\tilde{Y}$ has; negative $Z$ indicates that the caption misses some contents of the image (undercoverage), and positive $Z$ indicates that the caption includes something that is not in the image (hallucination).

\subsection{Objectives}
The image captioning process can be compared to a communication system~\cite{Shannon1948mathematical} where information source $X$ is converted to signal $\tilde{X}$ by a lossless source encoder $g$, transmitted through a noisy channel $f_\theta$, and the received signal $\tilde{Y}$ is losslessly decoded by $h$, giving the final received information $Y$. 
From this communication system perspective, one might say that the optimal $\theta^*$ could just be the one that minimizes the error $||Z||$. However, this requirement is unrealistic as it would result in extremely long captions that losslessly encode both high-level semantic information and all the low-level irrelevant information.

Therefore, we apply the information bottleneck principle~\cite{Tishby2000information,ShwartzZiv2017Opening,Tsai2021Self} to evaluate this system. Information bottleneck is a generalization of rate-distortion theory for lossy data compression, 
it requires a representation (which in our case is $\tilde{Y}$) to have maximal mutual information with some information $T$ that is required fulfill the task requirements (\textit{i.e.}, high information sufficiency),
while having minimal mutual information with the input $X$ (minimal redundancy). The desired minimal sufficient representation can be given as $\tilde{Y}^*=\arg\max I(\tilde{Y};T) - \beta I(X;\tilde{Y})$ where $\beta$ is a Lagrange multiplier. If the captioning model is deterministic (given $X$, it always produces the same $\tilde{Y}$) then we have $H(\tilde{Y}|X)=0$. Since $I(X;\tilde{Y})=H(\tilde{Y})-H(\tilde{Y}|X)$, the minimal sufficient representation can be written as:

\begin{equation}\label{eq:minimal sufficient representation}
    \tilde{Y}^* = \arg \underset{\tilde{Y}}{\max}\ I(\tilde{Y};T) - \beta H(\tilde{Y}).
\end{equation}

The second term will penalize the captioning model when it generates an over-length caption and the value of $\beta$ controls the penalty strength. Combined with the first term, the objective requires the model to preserve as much useful information as possible for the task, while keeping the captions as succinct as possible. 

Next, we give a formal definition of the information sufficiency objective of image captioning with importance of semantic units~\cite{Peyrard2019Simple}.

\begin{definition}\label{def:information sufficiency}
    \textbf{\upshape (Information Sufficiency)} For given $X$, let a latent variable $T \in \mathcal{S}$ represent the task-relevant information in $X$, and let an importance variable $A \in [0,1]^n$ denote the importance scores of different semantic units. The $A$ is derived from $X$ by an underlying mapping, thus being dependent on $X$. The $T$ is produced by a point-wise product between $A$ and $X$, thus $T=A\odot X$. For generated image captions $\tilde{Y}=f_\theta(\tilde{X})$, the information sufficiency objective is:
    \begin{equation}
        J_\mathrm{suf}(\theta)=I(\tilde{Y};A\odot X)
    \end{equation}
\end{definition}

In the importance variable, $A_i=1$ means the semantic units $w_i$ are very important in the image, while $A_i=0$ means $w_i$ is irrelevant. It behaves similarly to the attention mechanism~\cite{Bahdanau2015Neural}, which also produces a heatmap between zero to one according to the given input. Note that here the $A$ is not binary, the continual nature of $A$ gives a good property to $J_\mathrm{suf}(\theta)$: when the ``budget'' is limited (as there are also other objectives to optimize), more semantic units with higher importance score will be retained while less important ones will be discarded.

\begin{definition}\label{def:minimal redundancy}
    \textbf{\upshape (Minimal Redundancy)} The minimal redundancy objective encourages the image captioning model $f_\theta$ to eliminate irrelevant information, it is given by measuring the entropy of generated captions: 
    \begin{equation}
        J_\mathrm{min}(\theta)=-H(\tilde{Y}).
    \end{equation}    
\end{definition}

Combining $J_\mathrm{suf}(\theta)$ and $J_\mathrm{min}(\theta)$ ensures the captions are \textbf{minimal sufficient representations} of images. However, there is no guarantee that the generated captions can be understood by humans. Therefore, we need to measure the human comprehensibility of the captions using a third objective, which is the distributional similarity between $Y\mathrm{data}$ and natural language.

\begin{definition}\label{def:huaman interpretability}
    \textbf{\upshape (Human Comprehensibility)} Let $P_{\tilde{Y}}$ denote the probabilistic distribution of model-generated captions over $\mathcal{D_\mathrm{caption}}$, and let $P_\mathrm{lang}$ denote the distribution of human interpretable natural language. Given a certain statistical divergence measurement $D$, the human comprehensibility objective is:
    \begin{equation}
        J_\mathrm{int}(\theta)=-D(P_{\tilde{Y}} || P_\mathrm{lang}).
    \end{equation}    
\end{definition}

The overall objective of image captioning is a weighted combination of information sufficiency, minimal redundancy, and human comprehensibility:

\begin{equation}\label{eq:overall objective}
    J(\theta) = J_\mathrm{suf}(\theta) - \beta J_\mathrm{min}(\theta) - \gamma J_\mathrm{int}(\theta),
\end{equation}

where $\beta>0$ and $\gamma>0$ are weighting coefficients. 
Here, one of the factors that the coefficient $\beta$ in Eq.~\ref{eq:overall objective} controls is the length of generated captions. If we prefer more detailed, comprehensive captions, we have smaller $\beta$ and larger $\gamma$.


\section{Proposed Method: PoCa}
\label{Section 3 Pyramids}

\begin{figure}
    \centering
    \includegraphics[width=\linewidth]{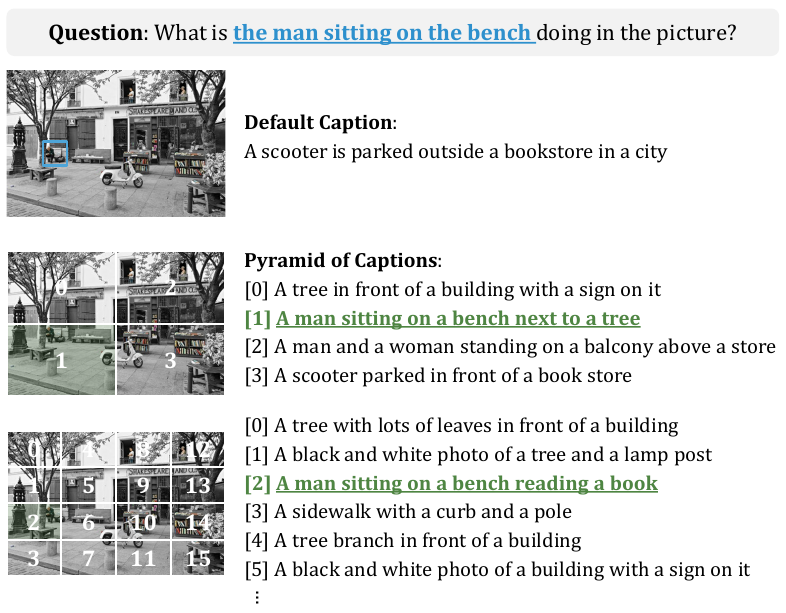}
    \caption{\label{fig:motivation_pyramid} Global caption tends to only capture the broad overview of the whole image. Local caption alleviates this limitation by providing fine-grained information which can further complement the global caption.}
\vspace{-10pt}
\end{figure}

In this section, we introduce the Pyramid of Captions (PoCa) method, which showcases the applicability of our framework to image captionoing research.
The key intuition behind the PoCa method is that we can have a more accurate and detailed caption by ensembling multiple captions, as shown in Figure~\ref{fig:motivation_pyramid}.
We propose to split an image into multiple local patches generate local captions for each patch, and fuse the captions to obtain a higher-quality caption for the global image.


Formally, let $\sigma_\mathrm{split}$ be a function operating in $\mathcal{D}_\mathrm{image}$ that represents the splitting function, which splits an image into a set of local patches:

\begin{equation}
\sigma_\mathrm{split}(\tilde{X})=\left\{\tilde{X}^{[j]} \right\}_{j=1}^m.
\end{equation}


We apply an image captioning model $f_\theta$ to the local patches and obtain a set of local captions: $\{\tilde{Y}^{[j]} = f_\theta(\tilde{X}^{[j]})\}_{j=1}^m$, and also generate a caption for the global image: $\tilde{Y} = f_\theta(\tilde{X})$. The local and global information will be fused by a merging function $\sigma_\mathrm{merge}$ operating in $\mathcal{D}_\mathrm{caption}$:

\begin{equation}
\tilde{Y}_\mathrm{merged} = \sigma_\mathrm{merge} \left( \tilde{Y}; \{\tilde{Y}^{[j]}\}_{j=1}^m \right).
\end{equation}

We adopt text-only LLMs as the merging function $\sigma_\mathrm{merge}$.
Table~\ref{tab:llm_caption_merging_prompt} in Appendix provides an example of merging using an LLM, where we instruct it to generate a merged caption that incorporates both local and global information.

As the PoCa method is hierarchical, we can extend it to be more layers. We can recursively split a patch into sub-patches and merge captions for each sub-patches to represent the patch. 


\section{PoCa Gives Better Captions (Provably)}
In this section, we provide an analysis of under what condition a single local-global merging operation in PoCa can be guaranteed to improve the caption quality. 


First, we assume that there is some function $\varphi$ to quantify the error $Z$ by $X$ in a deterministic manner, and that function is concave.
The deterministic assumption of $\varphi$ simplifies the analysis by ignoring errors caused by factors other than the input semantics $X$, such as the randomness in sampling-based autoregressive generation. In other words, we assume that $f_\theta$ always generates the same caption and makes the same error for the same input.
The concavity of $\varphi$ implies that it generates a larger volume of error when $p(w_i)$ is far from zero and one. This means that the captioning model is more likely to make mistakes when there is high uncertainty about the presence or absence of a semantic unit in the image. 

\begin{assumption}\label{assumption:Uncertainty-aware content-dependent error}
    \textbf{\upshape (Uncertainty-aware content-dependent error)} The error $Z$ produced by the image captioning model $f_\theta$ is dependent on the input semantics $X$. Therefore, it can be expressed as a deterministic function of $X$:
    \begin{equation}
        ||Z_i|| = \varphi(X_i),
    \end{equation}
    where $\varphi$ is a concave function and $1 \leq i \leq n$.
\end{assumption}

Next, we introduce our assumptions on the image splitting function $\sigma_\mathrm{split}$ and caption merging function $\sigma_\mathrm{merge}$. 
These assumptions simplify the relationship between local and global semantics by assuming linear combinations. In practice, the relationship may be more complex and depend on factors such as the spatial arrangement of the local patches and the presence of objects spanning multiple patches.

\begin{assumption}\label{assumption:Local-global relationship of image semantics}
    \textbf{\upshape (Local-global relationship of image semantics)} The $\sigma_\mathrm{split}$ splits an image into local patches. The latent semantic variables corresponding to local patches satisfy the following relationship with global semantics:
    \begin{equation}
        X = \sum_j^m \alpha_j X^{[j]},
    \end{equation}

where the weights $\alpha_j$ satisfying $\sum_j^m \alpha_j = 1$.
\end{assumption}

\begin{assumption}\label{assumption:Local-global aggregation of caption semantics}
    \textbf{\upshape (Local-global aggregation of caption semantics)}  The function $\sigma_\mathrm{merge}$ merges the global and global captions. The latent semantic variable corresponding to the merged caption is a weighted combination of the global and local semantics:
    \begin{equation}
        Y_\mathrm{merged} = \eta Y + (1-\eta) \sum_j^m \alpha_j Y^{[j]},
    \end{equation}
    where $\eta \in (0, 1)$ is a weighting coefficient.
\end{assumption}

We now present a theorem regarding $Z_\mathrm{merged} = Y_\mathrm{merged} - X$ (the proof can be found in the Appendix):

\begin{theorem}\label{theorem: poca is better}
    \textbf{\upshape (PoCa method reduces semantic error)}
    Under Assumptions 1-3, the PoCa method is guaranteed to have the smaller error $Z_\mathrm{merged}$ than $Z$, i.e.,

    \begin{equation}
    || Z_\mathrm{merged} || \leq || Z ||.
    \end{equation}

\end{theorem}

Since $A$ is non-negative, Theorem~\ref{theorem: poca is better} implies non-decreasing information sufficiency; if merging does not increase redundancy or decrease interpretability, then the overall quality of caption becomes better. This also aligns with the findings in~\cite{shi2024we} that smaller-scale models combined can be as effective as a larger-scale model. Additionally, it is worth noting that our assumptions require linear combinations of semantics, while it may not hold in practice for images with more complex structures of semantics.

\begin{proof}
First, we express the error of the $i$-th semantic unit in merged caption $Z_{\mathrm{merged},i}$ as the difference between the merged caption semantics $Y_{\mathrm{merged},i}$ and the source semantics $X_i$.
Using Assumption~\ref{assumption:Local-global aggregation of caption semantics} and \ref{assumption:Local-global relationship of image semantics}, $Z_{\mathrm{merged},i}$ can be expressed as:
{\small\begin{align}
    Z_{\mathrm{merged},i} &= Y_{\mathrm{merged},i} - X_i \\
    &= \eta Y_i + (1-\eta) \sum_j^m \alpha_j Y^{[j]}_i - X_i \label{eq:1_2} \\
    &= \eta (X_i + Z_i) + (1-\eta) \sum_j^m \alpha_j (X^{[j]}_i + Z^{[j]}_i) - X_i  \label{eq:1_3} \\
    &= \eta Z_i + (1-\eta) \sum_j^m \alpha_j Z^{[j]}_i.
\end{align}}

Here, (\ref{eq:1_3}) is derived from (\ref{eq:1_2}) is based on the decomposition of the global and local caption semantics into there corresponding source semantics and errors, i.e., $Y_i = X_i + Z_i$ and $Y^{[j]}_i = X^{[j]}_i + Z^{[j]}_i$.  

Next, we define $\Delta_{\mathrm{PoC},i}$ as the gap between the norm of the global caption error and the norm of the merged caption error for the $i$-th semantic unit. Using the triangle inequality and Assumption~\ref{assumption:Uncertainty-aware content-dependent error}, we derive a lower bound for $\Delta_{\mathrm{PoC},i}$:

{\small\begin{align}
\Delta_{\mathrm{PoC},i} &= ||Z_i|| - ||Z_{\mathrm{merged},i}|| \\
&= ||Z_i|| - ||\eta Z_i + (1-\eta) \sum_j^m \alpha_j Z^{[j]}_i|| \label{eq:2_2} \\
&\geq ||Z_i|| - (\eta ||Z_i|| + (1-\eta) \sum_j^m \alpha_j ||Z^{[j]}_i||) \label{eq:2_3} \\
&= (1-\eta) (||Z_i|| - \sum_j^m \alpha_j ||Z^{[j]}_i||) \\
&= (1-\eta) (\varphi(X_i) - \sum_j^m \alpha_j \varphi(X^{[j]}_i)). \label{eq:2_5}
\end{align}}

To yield (\ref{eq:2_3}) from (\ref{eq:2_2}), we apply the triangle inequality, which states that for any two vectors $a$ and $b$, $||a+b|| \leq ||a|| + ||b||$. 
Finally, we apply Assumption \ref{assumption:Uncertainty-aware content-dependent error} to obtain (\ref{eq:2_5}), which states $||Z_i|| = \varphi(X_i)$ and $||Z^{[j]}_i|| = \varphi(X^{[j]}_i)$.

By Assumption \ref{assumption:Uncertainty-aware content-dependent error}, $\varphi$ is a concave function. Applying Jensen's inequality and Assumption \ref{assumption:Local-global relationship of image semantics}, we have:

{\small\begin{equation}
\varphi(X_i) = \varphi(\sum_j^m \alpha_j X^{[j]}_i) \geq \sum_j^m \alpha_j \varphi(X^{[j]}_i).
\end{equation}}

This inequality implies that the error of the global caption is always greater than or equal to the weighted average of the errors of the local captions. Intuitively, this means that the PoCa method, which combines information from both global and local captions, is expected to have a lower error than using only the global caption.

Combining this result with the lower bound for $\Delta_{\mathrm{PoC},i}$ derived earlier, we can conclude that $\Delta_{\mathrm{PoC},i}$ is non-negative for all $i$:

{\small\begin{equation}
\Delta_{\mathrm{PoC},i} \geq (1-\eta) (\varphi(X_i) - \sum_j^m \alpha_j \varphi(X^{[j]}_i)) \geq 0.
\end{equation}}

The first inequality follows directly from the lower bound for $\Delta_{\mathrm{PoC},i}$ derived earlier. The second inequality follows from the Jensen's inequality result, which states that $\varphi(X_i) \geq \sum_j^m \alpha_j \varphi(X^{[j]}_i)$. Since $1-\eta > 0$ (as $\eta \in (0,1)$ by Assumption \ref{assumption:Local-global aggregation of caption semantics}), the product of $(1-\eta)$ and a non-negative term $(\varphi(X_i) - \sum_j^m \alpha_j \varphi(X^{[j]}_i))$ should also be non-negative, thus proving that $\Delta_{\mathrm{PoC},i} \geq 0$ for all $i$. Therefore, we have:

\begin{equation}
||Z_\mathrm{merged}|| \leq ||Z||.
\end{equation}

This completes the proof, demonstrating that under the given assumptions, the PoCa method is guaranteed to reduce the semantic error compared to using only the global caption. 

\end{proof}

\section{PoCa Gives Better Captions (Empirically)}
\subsection{Implementation Details}
\label{sec:Implementation Details}

\textbf{Image Captioning Models}. 
    We employ three groups of Large Vision Language Models (LVLMs) as the image captioning models: LLaVA-1.5~\cite{DBLP:journals/corr/abs-2310-03744}, MobileVLM v2~\cite{chu2024mobilevlm}, and InternVL~\cite{chen2023internvl}. Among them, \textbf{LLaVA-1.5} series is a popular LVLM with two variants: LLaVA-1.5-7B and LLaVA-1.5-13B, which adopt Vicuna-7B and Vicuna-13B as their Language Models (LLMs), respectively. \textbf{MobileVLM v2} is a family of efficient LVLMs with smaller scales, and we utilize its MobileVLM-v2-1.7B and MobileVLM-v2-3B models. \textbf{InternVL} is one of the top-performing publicly available LVLMs. We use its InternVL-Chat-Chinese-V1-2-Plus model, which is based on the Yi-34B LLM and has a total of 40.1B parameters.
    
    All inference is performed in FP16 precision on a single NVIDIA A800 GPU. We employ two types of prompts for short-form single-sentence image captioning and long-form detailed image captioning: \texttt{"Provide a one-sentence caption for the provided image"} and \texttt{"Describe this image in detail"}. All generation parameters are set to the default values provided by the source repository.

\textbf{Caption Pyramids}.
    For the caption merging function $\sigma_\mathrm{merge}$, we adopt and compare a variety of Large Language Models (LLMs) as its implementation, including the Gemma family (2B and 7B versions), LLaMA2 family (7B chat and 13B chat), Qwen-1.5-7B Chat, Mistral 7B, and a mixture-of-expert model Mixtral 8x7B. The Mixtral 8x7B model has capabilities similar to ChatGPT-3.5 and is one of the top-performing open-source LLMs. All inference is performed in FP16 precision, except for the large Mixtral 8x7B model, for which we use 8-bit quantization to fit it into a single NVIDIA A800 GPU. The Mixtral 8x7B is used as the default LLM for caption merging. 
    
    We employ the prompt shown in Table~\ref{tab:llm_caption_merging_prompt} for caption merging, where the \texttt{"Assistant Generation Prefix"} is injected after the instructions to control the model output format. All generation parameters are set to the default values provided by the source repository. For splitting function $\sigma_\mathrm{merge}$, we adopt the most straightforward implementation by splitting input image into four equal-sized patches.

\subsection{PoCa Improves Task Sufficiency}

\begin{table}[!t]
\begin{minipage}{\linewidth}\vspace{0mm}
\centering
\begin{tcolorbox}[colback=blue!0.8!white,colframe=blue!30!black,title=\textbf{Prompt for LLM-based VQA Evaluation}]
\centering
\small 
\begin{tabular}{p{\columnwidth} c}

    \VarSty{ {\bf System Message:} } & \\
        You will be given a caption of an image, and your task is to try to answer the question based on the caption.
        If the relevant information is not present in the caption, try your best to guess the answer.
        You shouldn't provide any rationale or explaination in your response, just give the answer only.
        The answer can be a number, a single word or a short phrase, plese make your response as short, simple and clear as possible.
        
        \hrulefill & \\
    
    \VarSty{ {\bf User:} } &\\
    
         Image Caption: \textcolor{red}{\texttt{\{image caption\}}} \> & \\
         Question: \textcolor{red}{\texttt{\{question\}}} \> & \\
        
        \hrulefill & \\
    
    \VarSty{ {\bf Assistant Generation Prefix:} } & \\
        The most possible answer is:& \\

\end{tabular}
\end{tcolorbox}
\caption{\textbf{Prompt for LLM-based VQA Evaluation}.
}
\label{tab:llm_vqa_evaluation_prompt}
\end{minipage}
\vspace{-10pt}
\end{table}

\subsubsection{VQA-based Evaluation}
    In this section, we conduct quantitative evaluations to study whether the PoCa method is able to improve the task sufficiency term in the objective (Definition~\ref{def:information sufficiency}). We adopt the VQA-v2~\cite{GoyalKSBP17} dataset, which is built upon the MS-COCO~\cite{ChenFLVGDZ15} dataset and contains multiple questions per image. 
    
    We employ a text-only LLM to answer the questions in VQA-v2 based on the generated captions, using the instruction shown in Table~\ref{tab:llm_vqa_evaluation_prompt}. The questions serve as a proxy for the importance score $A$ in Definition~\ref{def:information sufficiency}, and the accuracy of the LLM-generated correct answers becomes an estimation of the task sufficiency term. We use LLaMA2-Chat-13B for VQA inference; a detailed analysis of different LLMs for VQA evaluation will be provided in Section~\ref{sec:Impact of Different LLMs on VQA Evaluation}.

\begin{figure*}
    \centering
    \includegraphics[width=0.95\linewidth]{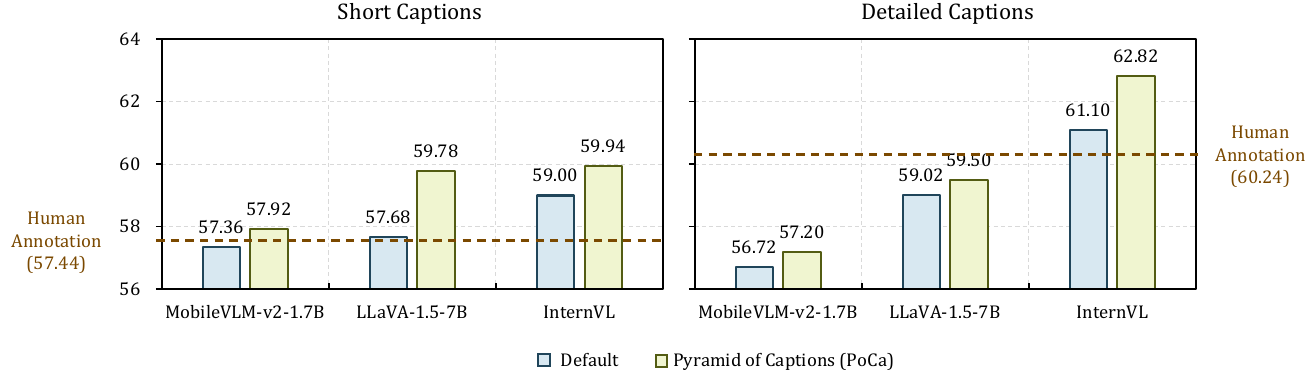}
    \caption{VQA accuracy using captions generated by different image captioning models and the proposed PoCa method. The PoCa method consistently improves the VQA performance across all models and caption types.}
    \label{fig:vqav2_main_results}
    \vspace{-5pt}
\end{figure*}

    Figure~\ref{fig:vqav2_main_results} provides the evaluation results on 5,000 questions in the VQA-v2 validation split. The human annotation for short captions represents the accuracy of a single-sentence caption drawn from the MS-COCO annotation, while the human annotation for detailed captions refers to five MS-COCO caption annotations concatenated with the prefix \texttt{"The following are several captions of this image written by different people: "} added to the front.
    
    As can be seen, our proposed PoCa method (green) consistently yields performance gains across all three examined LVLMs. The scale of improvement ranges from 0.48\% (MobileVLM-v2 detailed captions) to 2.10\% (LLaVA-1.5 short captions). Interestingly, we find that detailed captions do not necessarily correlate with better information coverage, as the detailed captions generated by MobileVLM-v2 underperform the single-sentence captions generated by InternVL. The comparison also shows that human annotations may not be optimal for certain scenarios, since several groups of LVLM-generated captions can yield higher VQA accuracy compared to that of human annotators.

\subsubsection{Image Paragraph Captioning}
    
    The image paragraph captioning dataset contains human-annotated single-paragraph descriptions for Visual Genome images. We use its testing split, which consists of 2,492 samples. We employ both the reference-free metric CLIPScore~\cite{HesselHFBC21} and the reference-based metric METEOR~\cite{BanerjeeL05} to evaluate the quality of the captions.
    
    CLIPScore measures the similarity between the image and text features extracted by the CLIP model (we use the standard OpenAI pretrained \texttt{ViT-Base-32}). The underlying assumption is that CLIP encoders are capable of extracting semantic information and can represent the importance score $A$; thus, a higher CLIPScore correlates with higher task sufficiency.
    
    The reference-based metric METEOR is widely adopted for evaluations in image captioning and natural language generation (\textit{e.g.}, machine translation). It measures the word-level similarity between model-generated captions and human-generated captions. The underlying assumption is that human annotations optimize the task sufficiency objective, so if a model behaves similarly to human annotations, it achieves high task sufficiency.
    
    The results are shown in Table~\ref{tab:image_paragraph_captioning}, where we also list the performance of previous fully-supervised models and few-shot models, including Regions-Hierarchical~\cite{krause2017hierarchical}, RTT-GAN~\cite{LiangHZGX17}, HSGED~\cite{YangGZ020}, SCST~\cite{Melas-KyriaziRH18}, CAE-LSTM~\cite{WangPYTM19}, VLIS~\cite{ChungY23}. Once again, our PoCa method provides task sufficiency improvement according to both the reference-free metric CLIPScore and the reference-based metric METEOR across all three families of LVLMs. These results further demonstrate the effectiveness of the PoCa method in enhancing the quality and informative content of the generated captions.

\begin{table}[]
\centering
\caption{Evaluation results on the image paragraph captioning dataset using CLIPScore and METEOR.}
\label{tab:image_paragraph_captioning}
\resizebox{\columnwidth}{!}{%
\begin{tabular}{ccccc}
\hline
\multicolumn{3}{c}{\textbf{Image Captioning Model}} & \textbf{CLIPScore} & \textbf{METEOR} \\ \hline
\multirow{5}{*}{\begin{tabular}[c]{@{}c@{}}Fully\\ Supervised \\ Models\end{tabular}} & \multicolumn{2}{c}{Regions-Hierarchical} & \multicolumn{1}{c}{-} & 15.95 \\
 & \multicolumn{2}{c}{RTT-GAN} & \multicolumn{1}{c}{-} & 17.12 \\
 & \multicolumn{2}{c}{HSGED} & \multicolumn{1}{c}{-} & 18.33 \\
 & \multicolumn{2}{c}{SCST} & \multicolumn{1}{c}{-} & 17.86 \\
 & \multicolumn{2}{c}{CAE-LSTM} & \multicolumn{1}{c}{-} & 18.82 \\ \hline
\multirow{14}{*}{\begin{tabular}[c]{@{}c@{}}Few-shot \&\\ Zero-shot\\ Models\end{tabular}} & BLIP-2 & 3-shot & \multicolumn{1}{c}{-} & 10.8 \\
 & OPT-IML & 3-shot & \multicolumn{1}{c}{-} & 9.5 \\
 & Naïve Ensemble & 3-shot & \multicolumn{1}{c}{-} & 9.8 \\
 & BLIP-2 & VLIS& \multicolumn{1}{c}{-} & 14.6 \\ \cline{2-5} 
 & \multirow{2}{*}{MobileVLM-v2-1.7B} & Default & 80.05 & 13.95 \\
 &  & PoCa & 81.80 & 16.39 \\ \cline{2-5} 
 & \multirow{2}{*}{MobileVLM-v2-3B} & Default & 79.02 & 8.99 \\
 &  & PoCa & 81.34 & 13.28 \\ \cline{2-5} 
 & \multirow{2}{*}{LLaVA-1.5-7B} & Default & 81.68 & 28.11 \\
 &  & PoCa & 81.80 & 28.79 \\ \cline{2-5} 
 & \multirow{2}{*}{LLaVA-1.5-13B} & Default & 82.16 & 28.44 \\
 &  & PoCa & 82.47 & 28.97 \\ \cline{2-5} 
 & \multirow{2}{*}{InternVL} & Default & 84.65 & 29.32 \\
 &  & PoCa & 85.52 & 29.84 \\ \hline
\end{tabular}%
}
\end{table}

\subsection{Ablation Study and Further Analysis}

\subsubsection{Caption Merging Strategies}
    In this section, we compare the effectiveness of different implementations of the merging function $\sigma_\mathrm{merge}$. First, we compare various LLMs introduced in Section~\ref{sec:Implementation Details}. As shown in Table~\ref{tab:Merging Strategy Ablations}, compared to the global caption baseline, every LLM yields performance improvement, except for the smallest Gemma-2B-IT model. We also provide an ablation on prompting, where we replace the default prompt shown in Table~\ref{tab:llm_caption_merging_prompt} with a naive prompt of \texttt{"merge these captions"}. This ablation results in a slight decrease in accuracy and a significant increase in caption length, which further violates the minimal redundancy objective.
    
    Additionally, we compare two parameter-free merging strategies based on simply concatenating local-only captions (representing $\eta=0$ in Assumption~\ref{assumption:Local-global aggregation of caption semantics}) or local-global captions, with positional encoding as in the \texttt{"User"} field in Table~\ref{tab:Merging Strategy Ablations}. The results show that local captions alone cannot provide sufficient information, while adding the global caption brings significant improvement. However, these two concatenation-based methods generate excessively long captions, demonstrating the necessity of LLM-based information fusion and length compression.

\begin{table}[]
\centering
\caption{Comparison of different caption merging strategies on the VQA-v2 validation set.}
\label{tab:Merging Strategy Ablations}
\resizebox{\columnwidth}{!}{%
\begin{tabular}{lrrr}
\hline
\textbf{Merging Function} & \textbf{Params} & \textbf{Accuracy} & \textbf{Length} \\ \hline
Global Caption Baseline & 0 & 57.68 & 50.75 \\ \hline
Gemma-2B-IT & 2B & 57.44 & 107.14 \\
Gemma-7B-IT & 7B & 58.74 & 178.79 \\
Mistral 7B Instruct-v0.2 & 7B & 58.92 & 136.12 \\
LLaMA2-7B Chat & 7B & 58.64 & 199.34 \\
LLaMA2-7B Chat (Naive Prompt) & 7B & 58.60 & 239.02 \\
Qwen-1.5-7B-Chat & 7B & 58.64 & 130.93 \\
LLaMA2 13B Chat & 13B & 59.06 & 154.67 \\
Mixtral 8x7B Instruct-v0.1 & 46.7B & 59.78 & 219.22 \\ \hline
Local Captions Concatenation & 0 & 55.66 & 265.63 \\
Global Local Concatenation & 0 & 59.12 & 337.38 \\ \hline
\end{tabular}%
}
\end{table}

\subsubsection{PoCa Does Not Sacrifices Minimal Redundancy}
    Previous evaluations primarily focused on the task sufficiency of image captions. However, as our objective also contains other terms, it is crucial to evaluate whether the performance gain brought by PoCa is achieved by significantly sacrificing other objectives. In Table~\ref{tab:caption_length}, we present the length statistics for captions used in the previous task sufficiency comparison. We calculate the average number of words in default captions and PoCa captions and note their differences in the ``$\pm\Delta$'' column. 
    
    The results show that PoCa does not exhibit a significant trend of either increasing or decreasing the length of captions. Among the six comparisons, PoCa compresses the length in four cases and extends the length in two cases. This empirically demonstrates that using LLMs as $\sigma_\mathrm{merge}$ in the PoCa model does not significantly violate the minimal redundancy objective $H(\tilde{Y}_\mathrm{merged})$.

\begin{table}[]
\centering
\caption{Comparison of caption lengths between default captions and PoCa captions on VQAv2 and Img2P datasets.}
\label{tab:caption_length}
\resizebox{\columnwidth}{!}{%
\begin{tabular}{lrrr|rrr}
\hline
\multirow{2}{*}{\textbf{LVLM}} & \multicolumn{3}{c|}{\textbf{VQAv2}} & \multicolumn{3}{c}{\textbf{Img2P}} \\
 & \textbf{Default} & \textbf{PoCA} & \textbf{$\pm\Delta$} & \textbf{Default} & \textbf{PoCA} & \textbf{$\pm\Delta$} \\ \hline
\multicolumn{1}{l}{MobileVLM-v2-1.7B} & 54.1 & 78.2 & +24.1 & 61.6 & 47.0 & -14.6 \\
LLaVA-1.5-7B & 82.7 & 74.7 & -8.0 & 93.2 & 133.4 & +40.2 \\
InternVL & 158.3 & 93.4 & -65.0 & 177.4 & 176.2 & -1.2 \\ \hline
\end{tabular}%
}
\end{table}



\section{Conclusion}
\label{Section 6 Conclusion}
Our work presents a novel information-theoretic framework that provides well-defined principles for image captioning covering information sufficiency, minimal redundancy, and human interpretability. By leveraging the theoretical framework, we propose Pyramid of Captions (PoCa), a novel image captioning approach that employs a hierarchical method to generate content-rich captions by exploiting the complementary nature of local and global visual cues. Through theoretical proofs and empirical evaluations, we demonstrate that PoCa consistently enhances the quality of image captions, making them more informative, semantically accurate, and contextually coherent while maintaining brevity and interoperability. 

\newpage

\section*{Limitations}

While the PoCa method has demonstrated effectiveness in improving image caption quality, there are several limitations that are worth discussing. 

\textbf{Assumptions on Image Semantics.}
    The Assumption~\ref{assumption:Local-global relationship of image semantics} made in this work could be sometimes strong and unrealistic, especially for the naive patch splitting function. The linear combination assumption may not hold well for images with more complex structures. This issue could be particularly problematic when objects or important semantic elements span across multiple local patches. In future work, employing more advanced splitting functions, object detection or semantic segmentation, could help alleviate this limitation and better capture the semantic structure of the image.
    
\textbf{Assumptions on Caption Semantics.}
    Similarly, the assumption about the local-global aggregation of caption semantics (Assumption~\ref{assumption:Local-global aggregation of caption semantics}) may not always be well satisfied by the LLM used for caption merging, particularly when the LLM is not sufficiently powerful. Weaker LLMs may struggle to effectively combine the local and global caption semantics in the desired manner. Further investigation into the impact of LLM choice on the fulfillment of this assumption would be valuable.
    
\textbf{Depth of the Caption Pyramid.}
    In the experiments, this work has demonstrated the benefits of a single level of local-global splitting and merging. However, the potential of deeper caption pyramids has not been fully explored. As the pyramid grows deeper, there could be a distribution shift for the input image patches, leading to more errors in the generated captions. Investigating the performance of merging functions for noisier captions is an important direction for future research.
    
\textbf{VQA Evaluation.}
    While the VQA-based evaluation provides a useful measure of caption quality in terms of information sufficiency, it has limitations. The questions used for evaluation may not comprehensively cover all of the important semantic units, resulting a sub-optimal estimation of the importance score $A$. In addition, due to resource constraints, we use a 5,000 question subset from the full VQAv2 dataset. To test its reliability, we run default caption generation with 5 models, together with human annotated caption, resulting in a total of (5+1)$\times$2=12 data points combining short and long captioierns. The Pearson correlation coefficient between 5k subset accuracy and full dataset accuracy is 0.8519 -- although already quite high, it still introduce some degree of noise for model performance evaluation.

\textbf{Computational Efficiency.} 
    Our implementation of PoCa involves more inferences to generate captions and prompting LLM for fusing the local and global captions. These multiple inference steps and the use of large models can lead to increased computational costs. This computational overhead may be a concern, especially in resource-constrained environments or when processing a large number of images. One potential solution is to finetune an image captioning model on the captions generated by PoCa. By doing so, the knowledge captured by PoCa can be distilled into the finetuned model, allowing for a single inference pass during deployment, while still benefiting from the enhanced caption quality achieved by PoCa. Similar approach of knowledge distillation has been adopted in other literature, such as~\cite{chen2023sharegpt4v,FanKIKT23,ChenLDW24}.

{\normalem
\bibliography{custom}
}

\newpage
\onecolumn
\appendix

\section*{\centering\textbf{Appendix}}

\begin{table*}[h!]
\begin{minipage}{\linewidth}\vspace{0mm}
\centering
\begin{tcolorbox}[colback=blue!0.9!white,colframe=blue!30!black,title=\textbf{Prompt for Merging Caption Pyramid}]
\centering
\footnotesize 
\begin{tabular}{p{\columnwidth} c}

    \VarSty{ {\bf System Message:} } & \\
        \textbf{Input}:& \\
            $\quad\bullet\quad$ You will receive a \textbf{global caption} describing an image.& \\
            $\quad\bullet\quad$ Additionally, you will have access to \textbf{local captions} generated for specific patches within the image. & \\
            $\quad\bullet\quad$ Both global and local captions may contain noise or errors.& \\
        \\
        \textbf{Task Objective}:& \\
            $\quad\bullet\quad$  Your goal is to create a \textbf{merged global caption} that combines relevant information from both sources.& \\
            $\quad\bullet\quad$  The merged caption should be \textbf{no longer than the original ones}.& \\
            $\quad\bullet\quad$  You only give the merged caption as output, \textbf{without any additional information}.& \\
            $\quad\bullet\quad$  Do NOT give any explaination or notes on how you generate this caption.& \\
        \\
        \textbf{Guidelines}:& \\
            $\quad\bullet\quad$  \textbf{Combine Information}: Extract key details from both global and local captions.& \\
            $\quad\bullet\quad$  \textbf{Filter Noise}: Remove non-sense content, inaccuracies, and irrelevant information.& \\
            $\quad\bullet\quad$  \textbf{Prioritize Visual Details}: Highlight essential visual elements instead of feeling or atmosphere& \\
            $\quad\bullet\quad$  \textbf{Be Concise}: Use as few words as possible while maintaining coherence and clarity.& \\
            $\quad\bullet\quad$  \textbf{Ensure Coherence}: Arrange the merged information logically.& \\
        \\
        Remember, your output should be a high-quality caption that is concise, informative, and coherent!& \\
        
        \hrulefill & \\
    
    \VarSty{ {\bf User:} } &\\
    
         \#\#\# Global Caption: \textcolor{red}{\texttt{\{global caption\}}} \> & \\
         \#\#\# Top-left: \textcolor{red}{\texttt{\{top-left caption\}}} \> & \\
         \#\#\# Bottom-left: \textcolor{red}{\texttt{\{bottom-left caption\}}} \> & \\
         \#\#\# Top-right: \textcolor{red}{\texttt{\{top-right caption\}}} \> & \\
         \#\#\# Bottom-left: \textcolor{red}{\texttt{\{bottom-left caption\}}} \> & \\
        
        \hrulefill & \\
    
    \VarSty{ {\bf Assistant Generation Prefix:} } & \\
        Here's the merged caption:& \\

\end{tabular}
\end{tcolorbox}
\caption{\label{tab:llm_caption_merging_prompt}An Example implementation of the merging function $\sigma_\mathrm{merge}$ based on prompting text-only LLMs.}

\end{minipage}
\end{table*}

\section{Related Work}
\label{Section 5 Related Work}

\subsection{Image Captioning}

    Image captioning lies at the intersection of computer vision and natural language processing, requiring both accurate visual recognition and coherent language generation abilities. In Section~\ref{Section 2 Framework} we formally defined the objectives for this task, but it is worth noting that current efforts do not explicitly optimize that objective. The primary challenge is the difficulty of back-propagation through the discrete textual space $\mathcal{D}_\mathrm{caption}$, and efforts addressing this challenge involve adopting reinforcement learning~\cite{abs-2205-12630, Feng00L19a} or aligning continuous latent spaces with language spaces~\cite{LiuYA23, YuCWKMHREB0MH023}. However, these approaches suffer from training instability and less satisfactory language coherence.
    
    Most current methods rely on a surrogate methodology, where they use human annotators $f_\mathrm{human}$ to write captions and train models to imitate those captions. The underlying assumption is that human-written captions optimize the objective $J(\mathrm{human})$, which is achieved by providing instructions to crowd-sourced caption annotators. For example, the instructions for MS COCO Caption annotation~\cite{ChenFLVGDZ15} include ``Describe all the important parts of the scene'' and ``Do not describe unimportant details'', which are respectively connected to the information sufficiency term and minimal redundancy term in our objective.
    
    An important trend in the image captioning field is the increasing focus on the comprehensiveness of image captions. As mentioned earlier, this represents a decreased length penalty (smaller weight $\beta$ for the minimal redundancy term) and more emphasis on the information sufficiency term. In recent years, there has been an increasing number of high-quality detailed captioning datasets for this target, such as human-annotated image paragraph captioning~\cite{krause2017hierarchical}, Densely Captioned Images (DCI)~\cite{urbanek2023picture}, and pseudo-labeled datasets, including LLaVA-Detailed-Captions~\cite{liu2023visual}, ShareGPT4V~\cite{chen2023sharegpt4v}, AS-1B~\cite{wang2024asm}, etc. However, detailed caption annotation is much more expensive than previous single-sentence annotation, while automated caption labeling exhibits a high risk of hallucination.

\subsection{Vision-Language Learning in the Era of Large Language Models}

    Various methods have been explored for enabling vision-language learning in LLMs. One line of work focusing on vision-language alignment during pretraining~\cite{dosovitskiy2020image,chen2020uniter,zhang2021vinvl,wang2022simvlm,wang2022ofa,yu2022coca,wang2022git,radford2021learning,li2022blip,li2023blip}, allowing the model to jointly learn a shared vision-language latent space. The other line of work, improve the vision-language training efficiency by aligning the vision representation into the language space of LLMs by only training the visual encoder module or a vision-language projection matrix~\cite{tsimpoukelli2021multimodal,alayrac2022flamingo,liu2023visual,sun2023emu,bai2023qwen,zhang2024llavar}. These two lines of works enable vision-language alignment, enabling various joint vision and language modalities prompting methods such as robot manipulation prompting~\cite{jiang2023vima,wake2023gpt} and multimodal in-context learning~\cite{zhao2024mmicl,zhang2024vicl}. 
    
    Unlike the other two directions, another line of work exploits the reasoning and planning ability of LLMs allowing zero-shot multimodal vision-language inference by extracting the information from the vision modality into a textual description and performing inference through frozen LLMs~\cite{yang2022pica,hu2022promptcap,lin2022revive,gui2022kat}. Recent works in this direction showcase remarkable VQA ability through answer heuristics generation~\cite{shao2023vllm} and enabling object 
    tagging and image editing through visual programming~\cite{gupta2023visual}. Inspired by this line of work, our work introduces a zero-shot hierarchical image captioning approach that relies on the reasoning ability of LLMs to aggregate information from local and global captions.

\section{VQA-based Caption Evaluation}
\label{sec:Impact of Different LLMs on VQA Evaluation}

    In this paper, we adopt the methodology of employing text-only LLMs for VQA inference with image captioning input to evaluate caption quality. This section provides a detailed analysis of this approach. Using the instruction given in Table~\ref{tab:llm_vqa_evaluation_prompt}, we prompt different LLMs to generate answers and evaluate the accuracy based on exact matching and Natural Language Inference (NLI) based evaluation. The NLI evaluation classifies a pair of statements, \texttt{"The answer to this question is \{ground truth\}"} and \texttt{"The answer to this question is \{generated answer\}"}, into entailment, neutral, and contradiction, where entailment outputs are regarded as successful. Compared to exact matching, NLI evaluation measures the correctness of answers at a semantic level and can tolerate low-level differences.
    
    As shown in Table~\ref{tab:llm_vqa_evaluation_prompt}, we find that different LLMs behave very differently in terms of answer length, and many LLMs fail to keep the answer succinct as instructed. Since the ground truth answers are mostly one word or a short phrase, this results in significantly reduced exact matching accuracy, while the actual semantic similarity is much higher, as measured by the NLI accuracy. We also observe an increasing trend in NLI accuracy when comparing different scales of LLMs, despite the largest Mixtral 8x7B Instruct-v0.1 producing lower NLI accuracy. We found that this outlier is caused by the over-conservative nature of the Mixtral 8x7B Instruct-v0.1 model, which frequently refuses to answer questions with responses such as ``cannot determine'' and ``not sure''. Finally, we add an ablation by instructing the LLM to guess the answer without caption input using the prompt: \texttt{"You will be given a question regarding an image, and your task is to try to infer the most possible answer"}. The resulting performance, noted as ``LLaMA2 7B Chat (No Caption)'', is much lower when measured by both exact matching and NLI accuracy.

\begin{table}[]
\centering
\caption{Comparison of different LLMs for VQA-based caption evaluation.}
\label{tab:my-table}
\resizebox{0.5\columnwidth}{!}{%
\begin{tabular}{lrrr}
\hline
\textbf{LLM} & \textbf{\begin{tabular}[c]{@{}c@{}}Answer\\ Length\end{tabular}} & \textbf{\begin{tabular}[c]{@{}c@{}}Match\\ Accuracy\end{tabular}} & \textbf{\begin{tabular}[c]{@{}c@{}}NLI\\ Accuracy\end{tabular}} \\ \hline
Gemma-2B-IT & 33.50 & 5.20 & 55.44 \\
Gemma-7B-IT & 38.00 & 0.00 & 54.44 \\
Mistral 7B Instruct-v0.2 & 28.90 & 2.30 & 63.30 \\
LLaMA2 7B Chat & 6.10 & 57.44 & 67.14 \\
LLaMA2 7B Chat (No Caption) & 4.30 & 41.34 & 44.76 \\
Qwen-1.5-7B-Chat & 7.90 & 56.72 & 69.06 \\
LLaMA2 13B Chat & 5.30 & 60.24 & 69.14 \\
Mixtral 8x7B Instruct-v0.1 & 24.40 & 8.38 & 64.86 \\ \hline
Ground Truth Answer & 4.70 & \multicolumn{1}{c}{-} & \multicolumn{1}{c}{-} \\ \hline
\end{tabular}%
}
\end{table}

\section{PoCa Examples}

    We show some example of PoCa caption merging, where the images are sourced from the test split of the image paragraph captioning dataset~\footnote{\url{https://cs.stanford.edu/people/ranjaykrishna/im2p/index.html} licensed by \href{https://homes.cs.washington.edu/~ranjay/visualgenome/about.html}{CC BY 4.0}.}. Captions are generated by LLaVA-1.5-7B model and merged by Mixtral-8x7B LLM. It shows that local captions successfully captured some errors that are hallucinated or omitted by the global caption, while PoCa successfully aggregated such information (marked by \textcolor{teal}{\uline{\textbf{green}}}) and recompile it into a human-interpretable caption.

\begin{table*}[h!]
\centering
\resizebox{\linewidth}{!}{
\begin{tabular}{p{3cm} l}
    \multirow{10}{*}{\includegraphics[width=3cm]{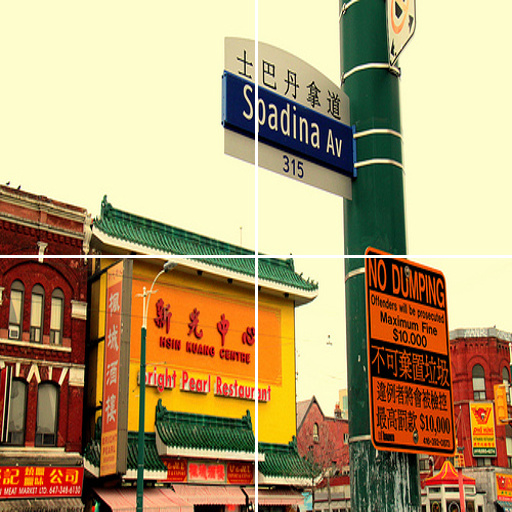}} & \begin{tabular}[t]{@{}m{\linewidth}@{}} \\
    \textbf{Global Caption} The image features a street corner with a street sign hanging above the \\
    street. The sign is written in both English and Chinese, indicating that the location is in an Asian \\
    neighborhood. The street sign is positioned above a restaurant, which is likely the "Asian Pearl \\
    Restaurant." There are several people in the scene, with some standing near the restaurant and \\
    others walking around the area. A car is parked on the street, and a traffic light is visible in the \\
    background. The overall atmosphere of the scene suggests a bustling urban environment with a mix of \\
    cultures. \\
     \\
    \end{tabular} \\
    \multirow{9}{*}{\includegraphics[width=3cm]{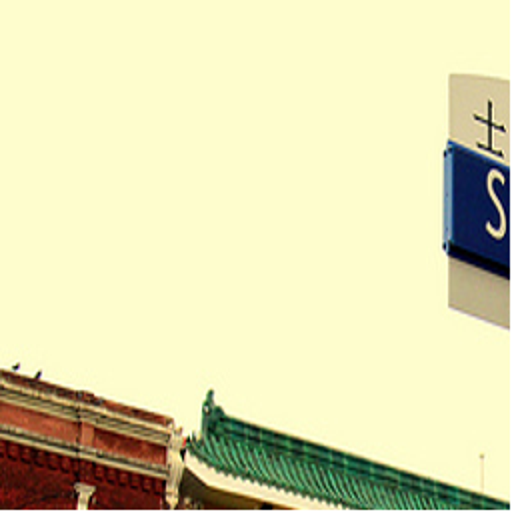}} & \begin{tabular}[t]{@{}m{\linewidth}@{}} \\
    \textbf{Top-left:} The image features a \textcolor{teal}{\uline{\textbf{large, ornate building with a green roof}}}. The building has \\
    a prominent green roof, which stands out against the sky. The sky is a pale yellow, creating a \\
    contrasting backdrop for the building. The building's architecture and the green roof make it an \\
    eye-catching structure in the scene. \\
     \\
     \\
     \\
    \end{tabular} \\
    \multirow{9}{*}{\includegraphics[width=3cm]{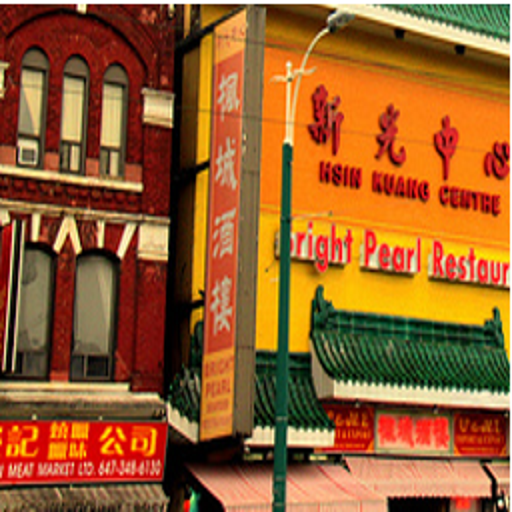}} & \begin{tabular}[t]{@{}m{\linewidth}@{}} \\
    \textbf{Bottom-left:} The image features a brightly colored building with a \textcolor{teal}{\uline{\textbf{yellow and red facade}}}, \\
    \textcolor{teal}{\uline{\textbf{likely a Chinese restaurant}}}. The building is adorned with a large sign that reads "Bright Pearl." \\
    The sign is positioned above the entrance, making it easily noticeable. In addition to the main \\
    building, there are two smaller buildings visible in the scene, one on the left side and the other \\
    on the right side. The overall atmosphere of the scene is vibrant and inviting. \\
     \\
     \\
    \end{tabular} \\
    \multirow{9}{*}{\includegraphics[width=3cm]{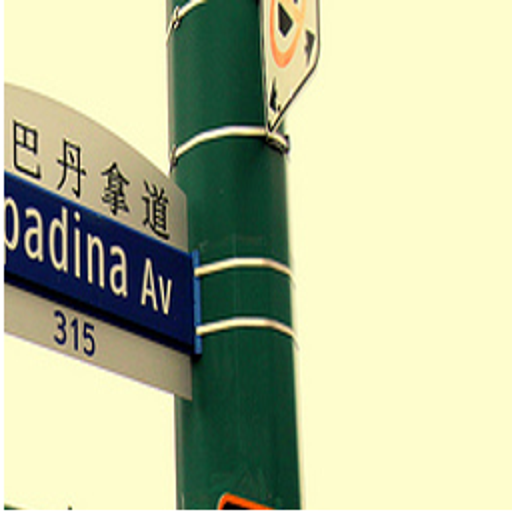}} & \begin{tabular}[t]{@{}m{\linewidth}@{}} \\
    \textbf{Top-right:} The image features a street sign with a blue background and white lettering, \\
    indicating the name of the street. The sign is attached to a green pole, which is located near a \\
    traffic light. The traffic light is positioned on the right side of the pole, and it is currently \\
    displaying a red light. The scene appears to be set in a city, with the street sign and traffic \\
    light being essential elements of urban infrastructure. \\
     \\
     \\
    \end{tabular} \\
    \multirow{9}{*}{\includegraphics[width=3cm]{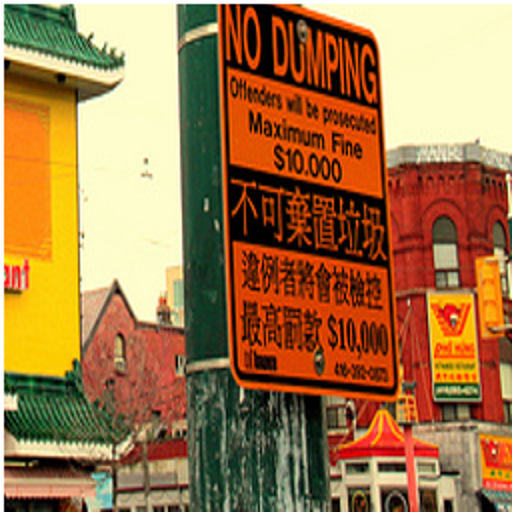}} & \begin{tabular}[t]{@{}m{\linewidth}@{}} \\
    \textbf{Bottom-right:} The image features a \textcolor{teal}{\uline{\textbf{street sign}}} in a foreign language, likely Chinese, \\
    \textcolor{teal}{\uline{\textbf{warning people about the consequences of dumping.}}} The sign is placed on a pole, and it is positioned \\
    near a building. The \textcolor{teal}{\uline{\textbf{sign is written in both English and Chinese}}}, making it accessible to a wider \\
    audience. In the background, there is a car parked on the street, and a person can be seen standing \\
    nearby. The scene appears to be set in an urban environment, with the street sign serving as a \\
    reminder for people to properly dispose of their waste. \\
     \\
    \end{tabular} \\
    \multirow{9}{*}{\includegraphics[width=3cm]{figures/VisualGenome_test_1255_Global.png}} & \begin{tabular}[t]{@{}m{\linewidth}@{}} \\
    \textbf{PoCa Merged Caption:} At a bustling Asian neighborhood street corner, an \textcolor{teal}{\uline{\textbf{ornate building}}} \\
    \textcolor{teal}{\uline{\textbf{with a green roof}}} and a \textcolor{teal}{\uline{\textbf{yellow-red Chinese restaurant}}} named "Bright Pearl" stand out. A street sign \\
    in English and Chinese indicates the location, while a \textcolor{teal}{\uline{\textbf{nearby sign warns against dumping}}}. A car, \\
    traffic light, and people add to the urban atmosphere. The likely "Asian Pearl Restaurant" is \\
    positioned near the street sign. \\
     \\
     \\
    \end{tabular} \\
    \multirow{9}{*}{} & \begin{tabular}[t]{@{}m{\linewidth}@{}} \\
    \textbf{Human Annotation:} A picture of a Asian town. There are many signs in an Asian language. Te \\
    sky above the town is gray. There is a large green pole to the right with signs mounted to it. There \\
    are multiple story buildings along the streets. There is a restaurant with the word pearl in it. \\
    Therre are blue and white signs on the pole. \\
     \\
     \\
     \\
    \end{tabular} \\
    
    \end{tabular}}
    
\end{table*}

\begin{table*}
\centering
\resizebox{\linewidth}{!}{
\begin{tabular}{p{3cm} l}
\multirow{9}{*}{\includegraphics[width=3cm]{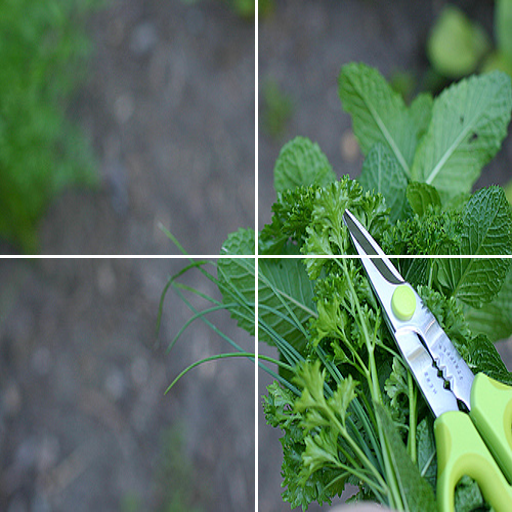}} & \begin{tabular}[t]{@{}m{\linewidth}@{}} \\
\textbf{Global Caption} The image features a person holding a pair of scissors with a leafy green \\
plant in their hand. The scissors are positioned in the middle of the plant, ready to cut it. The \\
person is likely preparing to trim the plant or remove a leaf. The scene takes place outdoors, with \\
the person standing in a garden or an outdoor area. \\
 \\
 \\
 \\
\end{tabular} \\
\multirow{9}{*}{\includegraphics[width=3cm]{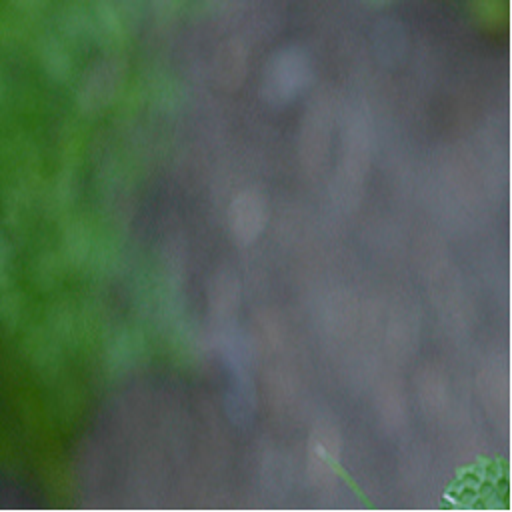}} & \begin{tabular}[t]{@{}m{\linewidth}@{}} \\
\textbf{Top-left:} The image features a close-up view of a patch of grass on a dirt ground. The \\
grass appears to be green and healthy, with a few \textcolor{teal}{\uline{\textbf{small rocks}}} scattered around the area. The \textcolor{teal}{\uline{\textbf{dirt}}} \\
\textcolor{teal}{\uline{\textbf{ground}}} is visible in the foreground, while the grassy patch extends into the background. The scene \\
captures the essence of a natural environment, with the grass and rocks creating a sense of depth \\
and texture. \\
 \\
 \\
\end{tabular} \\
\multirow{9}{*}{\includegraphics[width=3cm]{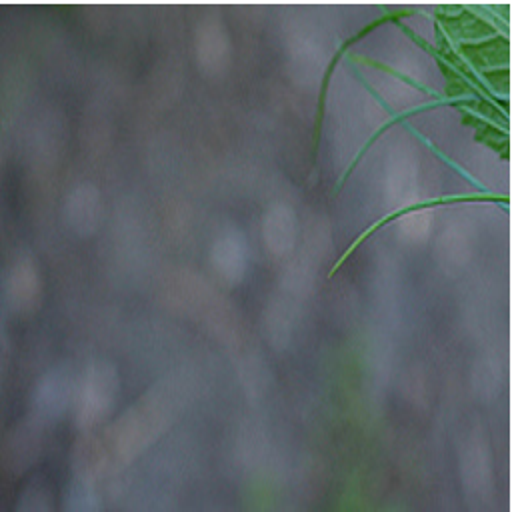}} & \begin{tabular}[t]{@{}m{\linewidth}@{}} \\
\textbf{Bottom-left:} The image features a close-up of a green plant with a few leaves. The plant \\
is situated in a dirt field, and its leaves are visible in the foreground. The plant appears to be \\
growing in a \textcolor{teal}{\uline{\textbf{dirt-covered area}}}, possibly a garden or a natural setting. The focus of the image is on \\
the plant and its leaves, creating a sense of depth and detail. \\
 \\
 \\
 \\
\end{tabular} \\
\multirow{9}{*}{\includegraphics[width=3cm]{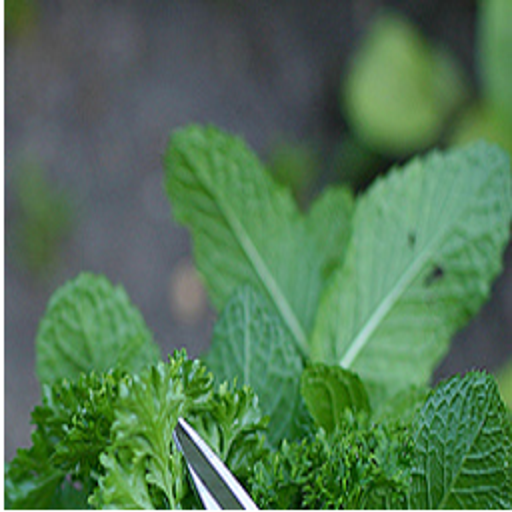}} & \begin{tabular}[t]{@{}m{\linewidth}@{}} \\
\textbf{Top-right:} The image features a close-up of a pair of scissors with a leafy green plant in \\
the foreground. The scissors are positioned in the middle of the scene, with the leafy plant \\
surrounding them. The plant appears to be a freshly cut herb, possibly \textcolor{teal}{\uline{\textbf{mint}}}, and is placed on a dirt \\
surface. The scissors are likely being used to trim the plant or prepare it for use in a recipe. \\
 \\
 \\
 \\
\end{tabular} \\
\multirow{9}{*}{\includegraphics[width=3cm]{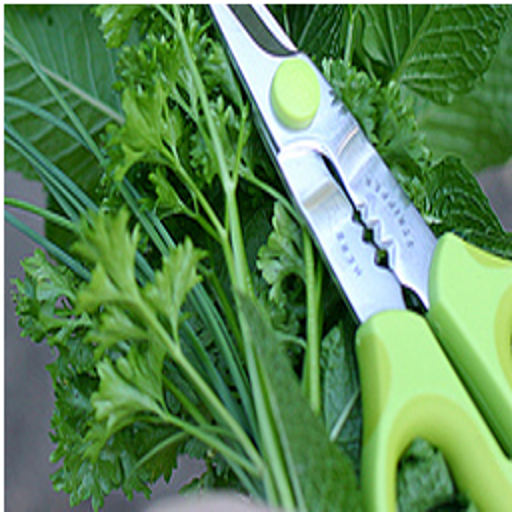}} & \begin{tabular}[t]{@{}m{\linewidth}@{}} \\
\textbf{Bottom-right:} The image features a pair of green scissors with a green handle, placed on a \\
pile of fresh herbs. The scissors are open, ready to be used for cutting the herbs. The herbs are \\
scattered around the scissors, with some located closer to the scissors and others further away. The \\
scene suggests that the person using the scissors is preparing to cut the herbs for cooking or other \\
purposes. \\
 \\
 \\
\end{tabular} \\
\multirow{9}{*}{\includegraphics[width=3cm]{figures/VisualGenome_test_223_Global.png}} & \begin{tabular}[t]{@{}m{\linewidth}@{}} \\
\textbf{PoCa Merged Caption:} A person holds scissors with a leafy green plant, likely preparing to \\
trim it in an outdoor setting. The scissors, situated in the middle of the plant, are positioned on \\
a pile of fresh herbs. The plant, possibly a type of \textcolor{teal}{\uline{\textbf{mint}}}, appears \textcolor{teal}{\uline{\textbf{healthy and green}}}, \textcolor{teal}{\uline{\textbf{surrounded by}}} \\
\textcolor{teal}{\uline{\textbf{small rocks and dirt}}}. \\
 \\
 \\
 \\
\end{tabular} \\
\multirow{9}{*}{} & \begin{tabular}[t]{@{}m{\linewidth}@{}} \\
\textbf{Human Annotation:} There are a pair of scissors sitting on top of a plant. The handle on \\
the scissors is colored green. The other part of the scissor is metal. The leaves of the plant or a \\
nice healthy green color. \\
 \\
 \\
 \\
 \\
\end{tabular} \\

\end{tabular}}

\end{table*}

\end{document}